\pdfoutput=1
\documentclass[11pt,a4paper]{article}
\usepackage{times,latexsym}
\usepackage{url}
\usepackage[T1]{fontenc}
\usepackage[utf8]{inputenc}
\usepackage{makecell}
\usepackage{multirow}
\usepackage{xcolor}
\usepackage{booktabs}

\usepackage{graphicx}
\usepackage[acceptedWithA]{tacl2018v2}
\usepackage{subcaption}

\usepackage{xspace,mfirstuc,tabulary}

\newif\iftaclinstructions
\taclinstructionsfalse % AUTHORS: do NOT set this to true
\iftaclinstructions
\renewcommand{\confidential}{}
\renewcommand{\anonsubtext}{(No author info supplied here, for consistency with
TACL-submission anonymization requirements)}
\newcommand{\instr}
\fi

\iftaclpubformat % this "if" is set by the choice of options

\else

\fi

\usepackage{xspace}

\newcommand{\GECCC}{\textit{GECCC}\xspace}
\newcommand{\GECCCLong}{\textit{Grammar Error Correction Corpus for Czech}\xspace}
\newcommand{\nsentences}{83\,058\xspace}

\newenvironment{citemize}{\begin{list}{$\bullet$}{\topsep=.1\smallskipamount\itemsep=0pt\parsep=1pt\labelwidth=.5em}}{\end{list}}

\title{Czech Grammar Error Correction with a Large and Diverse Corpus}

 \author{
 Jakub Náplava$^\dag$ \quad Milan Straka$^\dag$ \quad Jana Straková$^\dag$ \quad Alexandr Rosen$^\ddag$ \vspace{5.9pt}\\
 $^\dag$Charles University, Faculty of Mathematics and Physics \\
 Institute of Formal and Applied Linguistics \\
 \texttt{\{naplava,straka,strakova\}@ufal.mff.cuni.cz} \vspace{5.9pt}\\
 $^\ddag$Charles University, Faculty of Arts \\
 Institute of Theoretical and Computational Linguistics \\
 \texttt{alexandr.rosen@ff.cuni.cz}
}

\date{}

\usepackage[absolute]{textpos}
\begin{document}
\begin{textblock}{16}(0,0.1)\centerline{\small This paper was published in \textbf{TACL} -- please cite the published version \url{https://doi.org/10.1162/tacl_a_00470} instead.}\end{textblock}
\maketitle

\begin{abstract}

We introduce a large and diverse Czech corpus annotated for grammatical error correction (GEC) with the aim to contribute to the still scarce data resources in this domain for languages other than English. The \GECCCLong (\GECCC) offers a variety of four domains, covering error distributions ranging from high error density essays written by non-native speakers, to website texts, where errors are expected to be much less common. We compare several Czech GEC systems, including several Transformer-based ones, setting a strong baseline to future research. Finally, we meta-evaluate common GEC metrics against human judgements on our data. We make the new Czech GEC corpus publicly available under the CC BY-SA 4.0 license at {\footnotesize\url{http://hdl.handle.net/11234/1-4639}}.

\end{abstract}

\section{Introduction}

Representative data both in terms of size and domain coverage are vital for NLP systems development.  However, in the field of grammar error correction (GEC), most GEC corpora are limited to corrections of mistakes made by foreign or second language learners even in the case of English \cite{tajiri-etal-2012-tense,dahlmeier-etal-2013-building,yannakoudakis-etal-2011-new,yannakoudakis-2018-wi,ng-etal-2014-conll,napoles-etal-2017-jfleg}. At the same time, as recently pointed out by \citet{flachs-etal-2020-grammatical}, learner corpora are only a part of the full spectrum of GEC applications. To alleviate the skewed perspective, the authors released a corpus of website texts.

Despite recent efforts aimed to mitigate the notorious shortage of national GEC-annotated corpora \cite{boyd2018wnut,rozovskaya-roth-2019-grammar,davidson-etal-2020-developing-nlp,Ukrainian-GEC-2021,Romanian-GEC-2020,naplava-straka-2019}, the lack of adequate data is even more acute in languages other than English. 
We aim to address both the issue of scarcity of non-English data and the ubiquitous need for broad domain coverage by presenting a new, large and diverse Czech corpus, expertly annotated for GEC.

\GECCCLong (\GECCC) includes texts from multiple domains in a total of \nsentences sentences, being, to our knowledge, the largest non-English GEC corpus, as well as being one of the largest GEC corpora overall.

In order to represent a diversity of writing styles and origins, besides essays of both native and non-native speakers from Czech learner corpora, we also scraped website texts to complement the learner domain with supposedly lower error density texts, encompassing a representation of the following four domains:

\begin{citemize}
    \item \textit{Natives Formal} -- essays written by native students of elementary and secondary schools
    \item \textit{Natives Web Informal} -- informal website discussions
    \item \textit{Romani} -- essays written by  children and teenagers of the Romani ethnic minority
    \item \textit{Second Learners} -- essays written by non-native learners
\end{citemize}

Using the presented data, we compare several state-of-the-art Czech GEC systems, including some  Transformer-based.

Finally, we conduct a meta-evaluation of GEC metrics against human judgements to select the most appropriate metric for evaluating corrections on the new dataset. The analysis is performed across  domains, in line with \citet{napoles-etal-2019-enabling}.

Our contributions include (i) a \textbf{large and diverse Czech GEC corpus}, covering learner corpora and website texts, with unified and, in some domains, completely new GEC annotations, (ii) a \textbf{comparison of Czech GEC systems}, and (iii) a  \textbf{meta-evaluation of common GEC metrics} against human judgement on the released corpus.

\section{Related Work}

\subsection{Grammar Error Correction Corpora}

\begin{table*}[t]
    \centering
    \small
    \setlength{\tabcolsep}{3.3pt}
    \begin{tabular}{llrrlr}
        \toprule
        \multicolumn{1}{c}{Language} & \multicolumn{1}{c}{Corpus} & \multicolumn{1}{c}{Sentences} & \multicolumn{1}{c}{Err. r.} & \multicolumn{1}{c}{Domain} & \multicolumn{1}{c}{\makecell{\# Refs.}} \\
        \midrule
        \multirow{6}{*}{English} & \textit{Lang-8} & 1\,147\,451 &  14.1\% & SL & 1\\ %14.10
                                 & \textit{NUCLE} & 57\,151 & 6.6\%  & SL & 1\\ %6.56
                                 & \textit{FCE} & 33\,236  & 11.5\%  & SL  & 1 \\ %11.46
                                 & \textit{W\&I+LOCNESS} & 43\,169 & 11.8\%  & SL, native students& 5 \\ %11.8
                                 & \textit{CoNLL-2014 test}  & 1\,312 & 8.2\% & SL  & 2,10,8\\
                                 & \textit{JFLEG} & 1\,511 & --- & SL & 4\\
                                 & \textit{GMEG} & 6\,000 & --- & web, formal articles, SL & 4 \\
                                 & \textit{AESW} & over 1M & --- & scientific writing & 1\\
                                 & \textit{CWEB} & 13\,574 & $\sim$2\% & web & 2\\
        \midrule
        Czech & \textit{AKCES-GEC} & 47\,371 & 21.4\% & SL essays, Romani ethnolect of Czech & 2\\ %21.43
        \midrule
        German & \textit{Falko-MERLIN} & 24\,077 & 16.8\% & SL essays & 1\\ %16.84
        \midrule
        Russian & \textit{RULEC-GEC} & 12\,480 & 6.4\% & SL, heritage speakers & 1 \\
        \midrule
        Spanish &  \textit{COWS-L2H} & 12\,336 & --- & SL, heritage speakers & 2 \\
        \midrule
        Ukrainian & \textit{UA-GEC} & 20\,715 & 7.1\% & natives/SL, translations and personal texts & 2 \\
        \midrule
        Romanian & \textit{RONACC} & 10\,119 & --- & native speakers transcriptions & 1\\
        \bottomrule
    \end{tabular}
    \caption{Comparison of GEC corpora in size, token error rate, domain and number of reference annotations in the test portion. SL = second language learners.}
    \label{tab:data_overview}
\end{table*}

%Grammar error correction (GEC) corpora are predominantly based on essays written my mostly %non-native learners, with a few exceptions designed to broaden this high error density domain.
Until recently, attention has been focused mostly on English, while GEC data resources for other languages were in short supply. Here we list a few examples of English GEC corpora, collected mostly within an English-as-a-second-language (ESL) paradigm. For a comparison of their relevant statistics see Table~\ref{tab:data_overview}.

\textit{Lang-8 Corpus of Learner English} \cite{tajiri-etal-2012-tense} is a corpus of English language learner texts from the Lang-8 social networking system.

\textit{NUCLE} \cite{dahlmeier-etal-2013-building} consists of essays written by undergraduate students of the National University of Singapore. 

\textit{FCE} \cite{yannakoudakis-etal-2011-new} includes short essays written by non-native learners for the Cambridge ESOL First Certificate in English.

\textit{W\&I+LOCNESS} is a union of two datasets, the \textit{W\&I (Write \& Improve)} dataset \cite{yannakoudakis-2018-wi} of non-native learners' essays, complemented by the \textit{LOCNESS} corpus \cite{granger-1998}, a collection of essays written by native English students.

The GEC error annotations for the  learner corpora above were distributed with the BEA-2019 Shared Task on Grammatical Error Correction \cite{bryant-etal-2019-bea}.

The \textit{CoNLL-2014} shared task test set~\cite{ng-etal-2014-conll} is often used for GEC systems evaluation. This small corpus consists of 50 essays written by 25 South-East Asian undergraduates.

\textit{JFLEG} \cite{napoles-etal-2017-jfleg} is another frequently used GEC corpus with  fluency edits in addition to usual grammatical edits.

To broaden the restricted variety of domains, focused primarily on learner essays, a \textit{CWEB} collection \cite{flachs-etal-2020-grammatical} of website texts was recently released, aiming at contributing lower error density data.

\textit{AESW} \cite{daudaravicius-etal-2016-report} is a large corpus of scientific writing (over 1M sentences), edited by professional editors.

Finally, \citet{napoles-etal-2019-enabling} recently released \textit{GMEG}, a corpus for the evaluation of GEC metrics across domains.

Grammatical error correction corpora for languages other than English are less common and -- if available -- usually limited in size and domain: German \textit{Falko-MERLIN} \cite{boyd2018wnut}, Russian \textit{RULEC-GEC} \cite{rozovskaya-roth-2019-grammar}, Spanish \textit{COWS-L2H} \cite{davidson-etal-2020-developing-nlp}, Ukrainian \textit{UA-GEC} \cite{Ukrainian-GEC-2021} and Romanian \textit{RONACC} \cite{Romanian-GEC-2020}.

To better account for multiple correction options, datasets often contain several reference sentences for each original noisy sentence in the test set, proposed by multiple annotators. As we can see in Table~\ref{tab:data_overview}, the number of annotations typically ranges between 1 and 5 with an exception of the CoNLL14 test set, which -- on top of the official 2 reference corrections -- later received 10 annotations from ~\citet{bryant-ng-2015-far} and 8 alternative annotations from~\citet{sakaguchi2016reassessing}.

\subsection{Czech Learner Corpora}
\label{sec:learn-corp-czech}

% Jana: Dvojitý komentář - zkráceno zcela, jediný komentář: potenciálně možno vrátit, kdyby zbylo místo.

%%Plans for a broadly conceived learner corpus project of Czech went
%%ahead in 2009 with the collection of language written and spoken by
%%non-native and native learners.
By the early 2010s, Czech was one of a
few languages other than English to boast a series of learner corpora,
compiled under the umbrella project \textit{AKCES}, evoking the concept
of \textit{acquisition corpora} \cite{Sebesta:2010}.
%%The project,
%%funded since then on and off from different sources, turned out to be 
%%sustainable. New texts and recordings are collected, equipped with
%%better annotation, and available in more advanced search tools.
 
The native section includes transcripts of hand-written essays
(\textit{SKRIPT 2012}) and classroom conversation (\textit{SCHOLA
  2010}) from elementary and secondary schools. Both have their
counterparts documenting the Roma ethnolect of Czech:\footnote{The Romani ethnolect of Czech is the result of contact with Romani as the linguistic substrate. To a lesser (and weakening) extent the ethnolect shows some influence of Slovak or even Hungarian, because most of its speakers have roots in Slovakia. The ethnolect can exhibit various specifics across all linguistic levels. However, nearly all of them are complementary with their colloquial or standard Czech counterparts.  A short written text, devoid of phonological properties, may be hard to distinguish from texts written by learners without the Romani backround. The only striking exception are misspellings in contexts where the latter benefit from more exposure to written Czech. The typical example is the omission of word boundaries within phonological words, e.g. between a clitic and its host. In other respects, the pattern of error distribution in texts produced by ethnolect speakers is closer to native rather than foreign learners \cite{Borkovcova:2007,Borkovcova:2017}.} essays
(\textit{ROMi 2013}) and recordings and transcripts of dialogues
(\textit{ROMi 1.0}).\footnote{A more recent release \textit{SKRIPT
    2015} includes a balanced mix of essays from \textit{SKRIPT 2012}
  and \textit{ROMi 2013}. For more details and links see
  {\scriptsize\url{http://utkl.ff.cuni.cz/akces/}}.}
  
The non-native section goes by the name of \textit{CzeSL}, the acronym of
\textit{Czech as the Second Language}. \textit{CzeSL} consists of
transcripts of short hand-written essays %%(representing homework,
%%in-class or examination assignments)
collected from non-native
learners with various levels of proficiency and native languages, mostly students attending Czech language courses before or during their studies at a Czech university.
%%with a range of first language backgrounds and various levels
%%of proficiency.
%%Most of them are students attending Czech language
%%courses before or during their studies at a Czech university.
There
are several releases of \textit{CzeSL}, which differ mainly to what
extent and how the texts are annotated
\cite{Rosen:etal:2020}.\footnote{For a list of \textit{CzeSL} corpora with their sizes and annotation details see
  {\scriptsize\url{http://utkl.ff.cuni.cz/learncorp/}}.}

More recently, hand-written essays have been transcribed and annotated in \textit{TEITOK} \cite{Janssen:2016},\footnote{\scriptsize\url{http://www.teitok.org}} a tool combining a number of corpus compilation, annotation and exploitation functionalities.

Learner Czech is also represented in \textit{MERLIN}, a multilingual
(German, Italian and Czech) corpus built in 2012--2014 from texts
submitted as a part of tests for language proficiency levels
\cite{Boyd:etal:2014}.\footnote{\scriptsize\url{https://www.merlin-platform.eu}}
%The richly annotated, searchable and
%downloadable corpus is meant to illustrate the CEFR levels with examples of real
%language. Proficiency level indicated for each text is fairly reliable, because
%it is based on the test score.
%%On the other hand, the language tends
%%to be less inventive due to the primary goal of the examine to pass
%%the test.

Finally, \textit{AKCES-GEC} \cite{naplava-straka-2019} is a GEC corpus for Czech created from the subset of the above mentioned \textit{AKCES} resources \cite{Sebesta:2010}: the \textit{CzeSL-man} corpus (non-native Czech learners with manual annotation) and a part of the \textit{ROMi} corpus (speakers of the Romani ethnolect).

Compared to the \textit{AKCES-GEC}, the new \GECCC corpus contains much more data (47 371 sentences vs.\ \nsentences sentences, respectively), by extending data in the existing domains and also adding two new domains: essays written by native learners and website texts, making it the largest non-English GEC corpus and one of the largest GEC corpora overall.

\section{Annotation}

\subsection{Data Selection}

We draw the original uncorrected data from the following Czech learner corpora or Czech websites:

\begin{citemize}
    \item \textit{Natives Formal} -- essays written by native students of elementary and secondary schools from the \textit{SKRIPT 2012} learner corpus, compiled in the \textit{AKCES} project
    % language acquisition project 
    \item \textit{Natives Web Informal} -- newly annotated informal website discussions from Czech Facebook Dataset~\cite{data:czech_facebook,czech_facebook_dataset} and Czech news site \texttt{\footnotesize novinky.cz}. %\url{novinky.cz}
    \item \textit{Romani} -- essays written by children and teenagers of the Romani ethnic minority from the \textit{ROMi} corpus of the \textit{AKCES} project and the \textit{ROMi} section of the \textit{AKCES-GEC} corpus
    \item \textit{Second Learners} -- essays written by non-native learners, from the \textit{Foreigners} section of the \textit{AKCES-GEC} corpus, and the \textit{MERLIN} corpus
\end{citemize}

\begin{table}[t]
    \centering
    \small
    \begin{tabular}{lrr}
        \toprule
        Dataset & Documents & Selected \\
        \midrule
        \textit{AKCES-GEC-test} & 188 & 188 \\
        \textit{AKCES-GEC-dev} & 195 & 195 \\
        \textit{MERLIN} & 441 & 385 \\
        \textit{Novinky.cz} & --- & 2\,695 \\
        \textit{Facebook} & 10\,000 & 3\,850 \\
        \textit{SKRIPT\,2012} & 394 & 167 \\ 
        \textit{ROMi} & 1\,529 & 218\\
        \bottomrule
    \end{tabular}
    \caption{Data resources for the new Czech GEC corpus. The second column (Selected) shows the size of the selected subset from all available documents (first column, Documents).}
    \label{table:basic_dataset_domains}
\end{table}

\noindent
Since we draw our data from several Czech corpora originally created in different tools with different annotation schemes and instructions, we re-annotated the errors in a unified manner for the entire development and test set and partially also for the training set.

The data split was carefully designed to maintain representativeness, coverage and backwards compatibility. Specifically, (i) test and development data contain roughly the same amount of annotated data from all domains, (ii) original \textit{AKCES-GEC} dataset splits remain unchanged, (iii) additional available detailed annotations such as user proficiency level in \textit{MERLIN} were leveraged to support the split balance. Overall, the main objective was to achieve a representative cover over development and testing data. Table~\ref{table:basic_dataset_domains} presents the sizes of data resources in the number of documents. The first column (Documents) shows the number of all available documents collected in an initial scan. The second column (Selected) is a selected subset from the available documents, due to budgetary constraints and to achieve a representative sample over all domains and data portions. The relatively higher number of documents selected for the \textit{Natives Web Informal} domain is due to its substantially shorter texts, yielding fewer sentences; also, we needed to populate this part of the corpus as a completely new domain with no previously annotated data.

To achieve more fine-grained balancing of the splits, we used additional metadata where available: user's proficiency levels and origin language from \textit{MERLIN} and the age group from \textit{AKCES}.

%\AR{Jo, TEITOK jsme představili jako nástroj, ale tady není jasný, proč by nástroj měl umět age groups.}
% Jana: Z toho popisu v Related Work vyčnívá opravdu víc TEITOK než data, ale je to dáno tím, že TEITOK má referenci i url, kdežto data nemají. Ale věta alespoň začíná popisem těch dat.

\subsection{Preprocessing}

De/tokenization is an important part of data preprocessing in grammar error correction. Some formats, such as the M$^2$ format \cite{dahlmeier-ng-2012-better}, require tokenized formats to track and evaluate correction edits. On the other hand, detokenized text in its natural form is required for other applications. We therefore release our corpus in two formats: a tokenized M$^2$ format and detokenized format aligned at sentence, paragraph and document level. As part of our data is drawn from earlier, tokenized GEC corpora \textit{AKCES-GEC} and \textit{MERLIN}, this data had to be detokenized. A slightly modified Moses detokenizer\footnote{\scriptsize\url{https://github.com/moses-smt/mosesdecoder/blob/master/scripts/tokenizer/detokenizer.perl}} is attached to the corpus. To tokenize the data for the M$^2$ format, we use the UDPipe tokenizer \cite{UDPipe}.

% pro dev a test primocare

% pro train a Teitok v pohode; pro train a Merlin a AG nutno pouzit custom detokenizer (ten zverejnujeme taky)

%Train vsechno s puvodnimi starymi anotacemi: AG-train, Merlin rest, Tei rest

\subsection{Annotation}

The test and development sets in all domains were annotated from scratch
% with completely new annotations 
by five in-house expert annotators,\footnote{Our annotators are senior undergraduate students of humanities, regularly employed for various annotation efforts at our institute.} including re-annotations of the development and test data of the earlier GEC corpora to achieve a unified annotation style. All the test sentences were annotated by two annotators; one half of the development sentences received two annotations and the second half one annotation. The annotation process took about 350 hours in total. 

The annotation instructions were unified across all domains: The corrected text must not contain any grammatical or spelling errors and should sound fluent. Fluency edits are allowed if the original is incoherent. The entire document was given as a context for the annotation.
%and in case of the \textit{Natives Web Informal} domain, ten concatenated documents were annotated at once as the texts are rather short.
Annotators were instructed to remove too incomprehensible documents or those containing private information.

To keep the annotation process simple for the annotators, the sentences were annotated (corrected) in a text editor and postprocessed automatically to retrieve and categorize the GEC edits by the ERRor ANnotation Toolkit (ERRANT) \cite{bryant-etal-2017-automatic}.

% Jana: příliš detailní.
%- TODO v Novinkach pokud DELETE - tak delete (protoze private info)
%- nicmene v ostatnitch se obcas stane, ze jenom jeden DELETE (nesrozumitelne) - co pak?

%- z Teitoku filtrujeme vsechny dokumenty, co obsahuji annon (annonymized slovicko) kdekoliv v textu
%- z Melrinu filtujeme vsehcny dokumenty, co obsahuji unreadable kdekoliv v textu

\subsection{Train Data}

The first source for the training data are the data from the \textit{SKRIPT\,2012}; the \textit{MERLIN} corpus and the \textit{AKCES-GEC} train set that were not annotated, thus containing original annotations. These data cover the \textit{Natives Formal}, the \textit{Romani} and the \textit{Second Learners} domain. The second part of the training data are newly annotated data. Specifically, these are all \textit{Natives Web Informal} data and also a small part in the \textit{Second Learners} domain. All data in the training set were annotated with one annotation.

\subsection{Corpus Alignment}

The majority of models proposed for grammatical error correction operates over sentences. However, preliminary studies on document-level grammatical error correction recently appeared~\cite{chollampatt2019cross, yuan2021document}. The models were shown to benefit from larger context as certain errors such as errors in articles or tense choice do require larger context. To simplify future work with our dataset, we release three alignment levels: (i) sentence-level, (ii) paragraph-level and (iii) document-level. Given that the state-of-the-art grammatical error correction systems still operate on sentence level despite the initial attempts with document-level systems, we perform model training and evaluation at the usual sentence level.\footnote{Note that even if human evaluation in Section~\ref{sec:metrics} is performed on sentence-aligned data, human annotators process whole documents, and thus take the full context into account.}

\subsection{Inter-Annotator Agreement}

As suggested by \citet{rozovskaya-roth-2010-annotating}, followed later by \citet{rozovskaya-roth-2019-grammar} and \citet{Ukrainian-GEC-2021}, we evaluate  inter-annotator agreement by asking a second annotator to judge the need for a correction in a sentence already annotated by someone else, in a single-blind setting as to the status of the sentence (corrected/uncorrected).\footnote{A sentence-level agreement on sentence correctness is generally preferred in GEC annotations to an exact inter-annotator match on token edits, since different series of corrections may possibly lead to a correct sentence \cite{bryant-ng-2015-far}.} Five annotators annotated the first pass and three annotators judged the sentence correctness in the second pass. In the second pass, each of the three annotators judged a disjoint set of 120 sentences. Table~\ref{table:inter-annotator_agreement} summarizes the inter-annotator agreement based on second-pass judgements: the numbers represent the percentage of sentences judged correct in the second pass.

Both the average and the standard deviation ($82.96\pm12.12$) of our inter-annotator agreement are similar to inter-annotator agreement measured on English ($63\pm18.46$,  \citealt{rozovskaya-roth-2010-annotating}), Russian ($80\pm16.26$, \citealt{rozovskaya-roth-2019-grammar}) and Ukrainian ($69.5\pm7.78$ \citealt{Ukrainian-GEC-2021}).

\begin{table}[t]
    \centering
    \small
    \begin{tabular}{l|ccccc}
    \toprule
    \makecell[r]{First $\rightarrow$\\
    Second $\downarrow$} & A1 & A2 & A3 & A4 & A5 \\
    \midrule
    A1 & --- & 93.39 & 97.96 & 89.63 & 72.50 \\
    A2 & 84.43 & --- & 95.91 & 90.18 & 78.15 \\
    A3 & 68.80 & 87.68 & --- & 79.39 & 57.50 \\
    \bottomrule
    \end{tabular}
    \caption{Inter-annotator agreement based on second-pass judgements: numbers represent percentage of sentences judged correct in second-pass proofreading. Five annotators annotated the first pass, three annotators judged the sentence correctness in the second pass.}
    \label{table:inter-annotator_agreement}
\end{table}

\subsection{Error Type Analysis}
\label{ssec:errant}

To retrieve and categorize the correction edits from the erroneous-corrected sentence pairs, ERRor ANnotation Toolkit (ERRANT) \cite{bryant-etal-2017-automatic} was used. Inspired by \citet{boyd2018wnut}, we adapted the original English error types to the Czech language. For the resulting set see Table~\ref{table:errant_error_types}.  The POS error types are based on the UD POS tags \cite{nivre-etal-2020-universal} and may contain an optional \textit{:INFL} subtype when the original and the corrected words share a common lemma. The word-order error type was extended by an optional \textit{:SPELL} subtype to allow for capturing word order errors including words with minor spelling errors. The original orthography error type \textit{ORTH} covering both errors in casing and whitespaces is now subtyped with \textit{:WSPACE} and \textit{:CASING} to better distinguish between the two phenomena. Finally, we add two error types specific to Czech: \textit{DIACR} for errors in either missing or redundant diacritics and \textit{QUOTATION} for wrongly used quotation marks. Two original error types remain unchanged: \textit{MORPH}, indicating replacement of a token by another with the same lemma but different POS, and \textit{SPELL}, indicating incorrect spelling.

For part-of-speech tagging and lemmatization we rely on UDPipe \cite{UDPipe}.\footnote{Using the  czech-pdt-ud-2.5-191206.udpipe model.} The word list for detecting spelling errors comes from MorfFlex \cite{MorfFlexCZ20}.\footnote{We also use the \textit{aggresive} variant of the stemmer from {\scriptsize\url{https://research.variancia.com/czech_stemmer/}}.}

We release the Czech ERRANT at {\footnotesize\url{https://github.com/ufal/errant_czech}}. We assume that it is applicable to other languages with similar set of errors, especially Slavic languages, if lemmatizer, tagger and morphological dictionary are available.

\begin{table}[t]
    \centering
    \scriptsize
    \begin{tabular}{l|l|c}
    \toprule
     Error Type & Subtype & Example \\
    \midrule
    POS (15) & & \textit{tažené} $\rightarrow$ \textit{řízené} \\
    & :INFL & \textit{manželka} $\rightarrow$ \textit{manželkou} \\
    MORPH & & \textit{maj} $\rightarrow$ \textit{mají} \\
    ORTH & :CASING & \textit{usa} $\rightarrow$ \textit{USA} \\
     & :WSPACE & \textit{přes to} $\rightarrow$ \textit{přesto} \\
    SPELL & & \textit{ochtnat} $\rightarrow$ \textit{ochutnat} \\
    WO & & \textit{plná jsou} $\rightarrow$ \textit{jsou plná}\\
     & :SPELL & \textit{blískají zeleně} $\rightarrow$ \textit{zeleně blýskají} \\
    QUOTATION & & \textit{"} $\rightarrow$ \textit{„} \\
    DIACR & & \textit{tiskarna} $\rightarrow$ \textit{tiskárna} \\
    OTHER & & \textit{sem} $\rightarrow$ \textit{jsem ho}\\
    \bottomrule
    \end{tabular}
    \caption{Czech ERRANT Error Types.}
    \label{table:errant_error_types}
\end{table}

%TODO:
%- proteticke j (sem -> jsem; du -> jdu je ted OTHER mozna spise spell?)
%- krevetovej -> krevetovy je other

%\todo{JN: nevim, jestli to nechce ten ERRANT popsat trochu vic kdyz cilime na TACL?}
%Jana: Myslím si, že je popsaný docela podrobně.

\begin{figure*}[!t]
    %\leavevmode\kern-.1\hsize\includegraphics[width=1.2\hsize]{error_types_dev.pdf}
    \leavevmode\kern-.015\hsize\includegraphics[width=1.03\hsize]{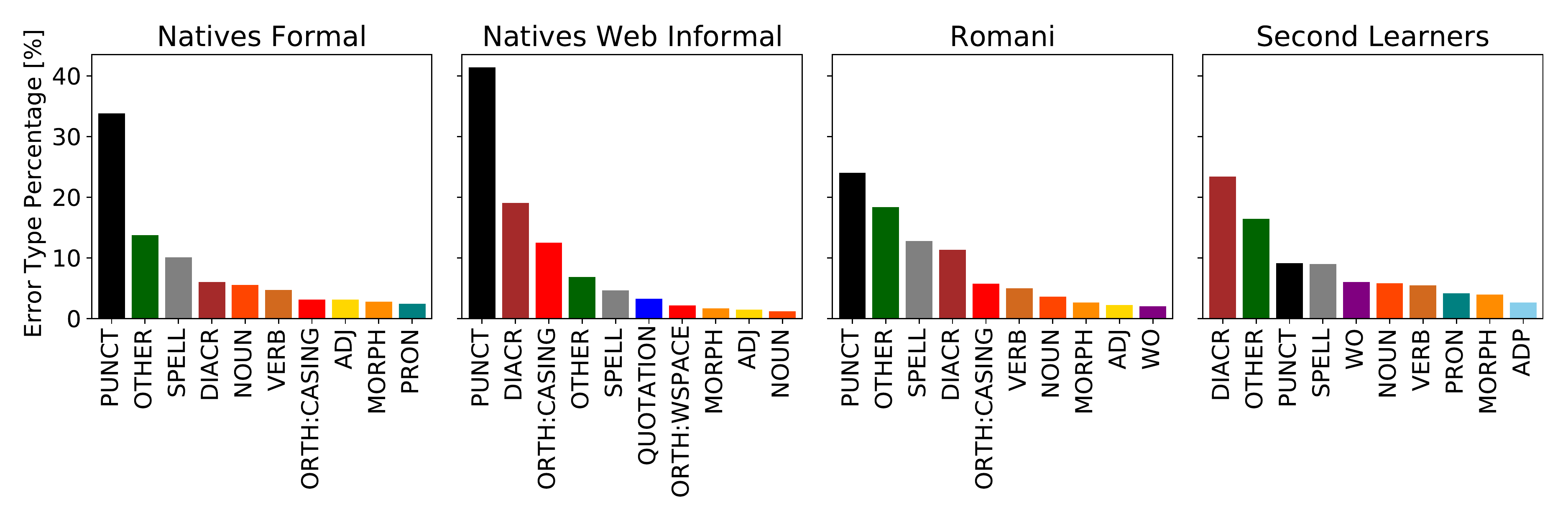}
    
    \vspace{-17pt}
    \caption{Distribution of top-10 ERRANT error types per domain in the development set.}
    \label{fig:errant_dataset_errors}
\end{figure*}

\begin{table*}[t]
    \centering
    \small
    \setlength{\tabcolsep}{5pt}
    \begin{tabular}{l|rrr|rrr|rrr|r}
    \toprule
     & \multicolumn{3}{c|}{Sentence-aligned} & \multicolumn{3}{c|}{Paragraph-aligned} & \multicolumn{3}{c|}{Doc-aligned} & \\
     & \multicolumn{3}{c|}{\#sentences} & \multicolumn{3}{c|}{\#paragraphs} & \multicolumn{3}{c|}{\#docs} & Error Rate \\
    & Train & Dev & Test & Train & Dev & Test & Train & Dev & Test & \\
    \midrule
    \textit{Natives Formal} & 4\,060 & 1\,952 & 1\,684 & 1\,618 & 859 & 669 & 227 & 87 & 76 & 5.81\% \\
    \textit{Natives Web Informal} & 6\,977 & 2\,465 & 2\,166 & 3\,622 & 1\,294 & 1\,256 & 3\,619 & 1\,291 & 1\,256 & 15.61\% \\
    \textit{Romani} & 24\,824 & 1\,254 & 1\,260 & 9\,723 & 574 & 561 & 3\,247 & 173 & 169 & 26.21\% \\
    \textit{Second Learners} & 30\,812 & 2\,807 & 2\,797 & 8\,781 & 865 & 756 & 2\,050 & 167 & 170 & 25.16\%\\
    \midrule
    Total & 66\,673 & 8\,478 & 7\,907 & 23\,744 & 3\,592 & 3\,242 & 9\,143 & 1\,718 & 1\,671 & 18.19\% \\
    \bottomrule
    \end{tabular}
    \caption{Corpus statistics at three alignment levels: sentence-aligned, paragraph-aligned and doc-aligned. Average Error rate was computed on the concatenation of development and test data in all three alignment levels.}
    \label{table:statistics_final_dataset}
\end{table*}

\subsection{Final Dataset}

The final corpus consists of \nsentences sentences and is distributed in two formats: the tokenized M$^2$ format \cite{dahlmeier-ng-2012-better} and the detokenized format with alignments at the sentence, paragraph and document levels. Although the detokenized format does not include correction edits, it does  retain full information about the original spacing.

\looseness1
The statistics of the final dataset are presented in Table~\ref{table:statistics_final_dataset}. The individual domains are balanced on the sentence level in the development and testing sets, each of them containing about 8\,000 sentences. The number of paragraphs and documents varies: on average, the \textit{Natives Web Informal} domain contains less than 2 sentences per document, while the \textit{Natives Formal} domain more than 20.

As expected, the domains differ also in the error rate, i.e.\ the proportion of erroneous tokens (see Table~\ref{table:statistics_final_dataset}). The students' essays in the \textit{Natives Formal} domain  are almost 3 times less erroneous than any other domain, while in the \textit{Romani} and \textit{Second Learners} domain, approximately each 4-th token is incorrect.

\looseness1
Furthermore, the prevalence of error types differs for each individual domain. %As described in Section~\ref{ssec:errant}, we enriched the data with error type annotations.
The 10 most common error types in each domain are presented in Figure~\ref{fig:errant_dataset_errors}. Overall, errors in punctuation (\textit{PUNCT}) constitute the most common error type. They are the most common error in three domains, although their relative frequency varies. We further estimated that of these errors, 9\% (\textit{Natives Formal}) -- 27\% (\textit{Natives Web Informal}) are uninteresting from the linguistic perspective, as they are only omissions of the sentence formal ending, probably purposeful in case of \textit{Natives Web Informal}. The rest (75--91\%) appears in a sentence, most of which (35--68\% \textit{Natives Formal}) is a misplaced comma: In Czech, syntactic status of finite clauses strictly determine the use of commas in the sentence. Finally, in 5--7\% cases of all punctuation errors, a correction included joining two sentences or splitting a sentence into two sentences. Errors in either missing or wrongly used diacritics (\textit{DIACR}), spelling errors (\textit{SPELL}) and errors in orthography (\textit{ORTH}) are also common, with varying frequency across domains.

Compared to the \textit{AKCES-GEC} corpus, the \GECCCLong contains more than 3 times as many sentences in the development and test sets, more than 50\%  sentences in the training set and also two new domains.

To the best of our knowledge, the newly introduced \GECCC dataset is the largest among GEC corpora in languages other than English and it is surpassed in size only by the English \textit{Lang-8} and \textit{AESW} datasets. 
%However, the annotations of Lang-8 are often of poor quality due to the fact that the texts are %corrected by online community users.
With the exclusion of these two datasets, the \GECCC dataset contains more sentences than any other GEC corpus currently known to us.

\section{Model}
\label{sec:model}

In this section, we describe five systems for automatic error correction in Czech and analyze their performance on the new dataset. Four of these systems represent previously published Czech work \cite{Richter:etal:2012,naplava-straka-2019,kazitext} and one is our new implementation. The first system is a pre-neural approach, published and available for Czech \cite{Richter:etal:2012}, included for historical reasons as a previously known and available Czech GEC tool; the following four systems represent the current state of the art in GEC: they are all neural network architectures based on Transformers, differing in the training procedure, training data or training objective. A comparison of systems, trained and evaluated on English, Czech, German and Russian, with state of the art is given in Table~\ref{table:english_german_russian_results}.

\begin{table*}
    \small
    \centering
    \begin{tabular}{l|c|cc|c|c|c}
        \toprule
        \multirow{2}{*}{System} & \multirow{2}{*}{Params} & \multicolumn{2}{c|}{English} & Czech & German & Russian \\
        & & W\&I+L & CoNLL 14 & AKCES-GEC & Falko-Merlin & RULEC-GEC \\
        \midrule
        %%% Best numbers setting
        \citet{boyd2018wnut} & -- &  -- & -- & -- & 45.22 & -- \\ % ok
        \citet{choe-etal-2019-neural} & -- & 63.05 & -- & -- & -- & -- \\ % ok
        \citet{lichtarge-etal-2019-corpora} & -- & -- & 56.8~~ & -- & -- \\ % ok
        \citet{lichtarge-etal-2020-data} & -- & 66.5~~ & 62.1~~ & -- & -- & -- \\ % ok
        \citet{omelianchuk-etal-2020-gector} & -- & 72.4~~ & 65.3~~ & -- & -- & -- \\ % ok
        \citet{rothe-etal-2021-simple} \textit{base} & 580M & 60.2~~ & 54.10 & 71.88 & 69.21 & 26.24 \\ % ok
        \citet{rothe-etal-2021-simple} \textit{xxl} & 13B & 69.83 & 65.65 & 83.15 & 75.96 & 51.62 \\ % ok
        \citet{rozovskaya-roth-2019-grammar} & -- & -- & -- & -- & -- & 21.00 \\ % ok
        \citet{xu-etal-2019-erroneous} & -- & 63.94 & 60.90 & -- & -- & -- \\ % ok
        \midrule
        \textit{AG finetuned} & 210M & 69.00 & 63.40 & 80.17 & 73.71 & 50.20 \\ % ok
        \bottomrule
    \end{tabular}
    \caption{Comparison of selected single-model systems on English (W\&I+L, CoNLL-2014), Czech (AKCES-GEC), German (Falko-Merlin GEC) and Russian (RULEC-GEC) datasets. Our reimplementation of the \textit{AG~finetuned} model is from \citet{naplava-straka-2019}. Note that models vastly differ in training/fine-tuning data and size (e.g., \citet{rothe-etal-2021-simple} \textit{xxl} is 50 times larger than \textit{AG~finetuned}).}
    \label{table:english_german_russian_results}
\end{table*}

\begin{table*}[t]
\small
\centering
\begin{tabular}{l|rrrrr|rrrrr}
\toprule
                      & \multicolumn{5}{c|}{$M^2_{0.5}$-score}
                      & \multicolumn{5}{c}{Mean human score}  
                      \\%\midrule
System                & \multicolumn{1}{c}{NF} & \multicolumn{1}{c}{NWI} & \multicolumn{1}{c}{R} & \multicolumn{1}{c}{SL} & \multicolumn{1}{c|}{$\Sigma$} 
                    & \multicolumn{1}{c}{NF} & \multicolumn{1}{c}{NWI} & \multicolumn{1}{c}{R} & \multicolumn{1}{c}{SL} & \multicolumn{1}{c}{$\Sigma$}
            \\\midrule
% Original & 8.47 & 7.99 & 7.76 & 7.18 & 7.61  & - & - & - & - \\\hline
% Korektor & 8.26 & 7.60 & 7.90 & 7.55 & 7.63& 26.47 & 24.38 & 36.53 & 43.41 & 35.63\\
% Synthetic trained & 8.55 & 7.99 & 8.10 & 7.88 & 7.98 & 33.59 & 25.87 & 31.63 & 47.74 & 38.37\\
% AG finetuned & 8.97 & 8.22 & 8.91 & 8.35 & 8.38 & 56.82 & TODO & 62.69 & 61.84 & TODO\\
% GECCC finetuned & --- & --- & --- & --- & --- & 65.52 & TODO & 65.61 & 66.10 & TODO \\
% Joint GEC+NMT & 9.06 & 8.37 & 8.69 & 8.19 & 8.35 & 59.20 & 53.69 & 54.66 & 59.14 & 56.05  \\\hline
% Reference & 9.58 & 9.48 & 9.60 & 9.63 & 9.57 & - & - & - & - \\
\textit{Original} & --- & --- & --- & --- & ---  & 8.47 & 7.99 & 7.76 & 7.18 & 7.61   \\
\midrule
\textit{Korektor} & 28.99 & 31.51 & 46.77	& 55.93 & 45.09 & 8.26 & 7.60 & 7.90 & 7.55 & 7.63 \\
\textit{Synthetic trained} & 46.83 & 38.63 & 46.36 & 62.20 & 53.07 & 8.55 & 7.99 & 8.10 & 7.88 & 7.98 \\
\textit{AG finetuned} & 65.77 & 55.20 & 69.71 & 71.41 & 68.08 & 8.97 & 8.22 & \textbf{8.91} & 8.35 & 8.38 \\
\textit{\textbf{GECCC finetuned}} & \textbf{72.50} & \textbf{71.09} & \textbf{72.23} & \textbf{73.21} & \textbf{72.96} & \textbf{9.19} & \textbf{8.72} & \textbf{8.91} & \textbf{8.67} & \textbf{8.74} \\
\textit{Joint GEC+NMT} & 68.14 & 66.64 & 65.21 & 70.43 & 67.40  & 9.06 & 8.37 & 8.69 & 8.19 & 8.35  \\
\midrule
\textit{Reference} & --- & --- & ---  & --- & --- & 9.58 & 9.48 & 9.60 & 9.63 & 9.57  \\
\bottomrule

\end{tabular}
\caption{Mean score of human judgements and $M^2_{0.5}$ score for each system in domains (NF = \textit{Natives Formal}, NWI = \textit{Natives Web Informal}, R = \textit{Romani}, SL = \textit{Second Learners}, $\Sigma$ = whole dataset). All results in the whole dataset (the $\Sigma$ column) are statistically significant with p-value $< 0.001$, except for the \textit{AG finetuned} and \textit{Joint GEC+NMT} systems, where the p-value is less than $6.2\%$ for $M^2_{0.5}$ score and less than $4.3\%$ for human score, using the Monte Carlo permutation test with 10M samples and probability of error at most $10^{-6}$ \cite{fay-mc-permutation-test,gandy-sequential-mc-risk}.}
\label{table:m2_human_scores}
\end{table*}

\subsection{Models}
\label{ssec:models}

We experiment with the following models:

\textit{\bfseries Korektor}~\cite{Richter:etal:2012} is a pre-neural statistical spellchecker and (occasional) grammar checker. It uses the noisy channel approach with a candidate model that for each word suggests its variants up to a predefined edit distance. Internally, a Hidden Markov Model \cite{baum1966statistical} is built. Its hidden states are the variants of words proposed by the candidate model, and the transition costs are determined from three $N$-gram language models built over word forms, lemmas and part-of-speech-tags. To find an optimal correction, Viterbi algorithm~\cite{forney1973viterbi} is used.

% Possibly Transformer synthetic pretrain
\textit{\bfseries Synthetic trained}~\cite{naplava-straka-2019} is a neural-based Transformer model that is trained to translate the original ungrammatical text to a well formed text. The original Transformer model~\cite{vaswani2017attention} is regularised with an additional source and target word dropout and the training objective is modified to focus on tokens that should change \cite{grundkiewicz-junczys-dowmunt-2019-minimally}. As the amount of existing annotated data is small, an unsupervised approach with a spelling dictionary is used to generate a large amount of synthetic training data. The model is trained solely on these synthetic data.

% Possibly Transformer finetune
\textit{\bfseries AKCES-GEC (AG) finetuned}~\cite{naplava-straka-2019} is based on \textit{Synthetic trained}, but finetunes its weights on a mixture of synthetic and authentic data from the \textit{AKCES-GEC} corpus, i.e., on data from the \textit{Romani} and \textit{Second Learners} domains. See Table~\ref{table:english_german_russian_results} for comparison with state of the art in English, Czech, German and Russian.

\textit{\bfseries GECCC finetuned} uses the same architecture as \textit{Synthetic trained}, but we finetune its weights on a mixture of synthetic and (much larger) authentic data from the newly released GECCC corpus. We use the official code of ~\citet{naplava-straka-2019} with the default settings and mix the synthetic and new authentic data in a ratio of 2:1.

\textit{\bfseries{Joint GEC+NMT}}~\cite{kazitext} is a Transformer model trained in a multi-task setting. It pursues two objectives: (i) to correct Czech and English texts; (ii) to translate the noised Czech texts into English texts and the noised English texts into Czech texts. The source data come from the \textit{CzEng v2.0} corpus~\cite{kocmi2020announcing} and were noised using a statistical system KaziText~\cite{kazitext} that tries to model several most frequently occurring errors such as diacritics, spelling or word ordering. The statistics of the Czech noise were estimated on the new training set, therefore, the system was indirectly trained also on data from \textit{Natives Formal} and \textit{Natives Web Informal} domains, unlike the \textit{AG finetuned} system. The statistics of the English noise were estimated on \textit{NUCLE} \cite{dahlmeier-etal-2013-building}, \textit{FCE} \cite{yannakoudakis-etal-2011-new} and \textit{W\&I+LOCNESS} \cite{yannakoudakis-2018-wi,granger-1998}.

\subsection{Results and Analysis}

Table~\ref{table:m2_human_scores} summarizes the evaluation of the five grammar error correction systems (described in the previous Section~\ref{ssec:models}), evaluated with highest-correlating and widely used metric, the $M^2$ score with $\beta=0.5$, denoted as $M^2_{0.5}$ (left); and with human judgements (right). For the meta-evaluation of GEC metrics against human judgements, see the following Section~\ref{sec:metrics}.

Clearly, learning on GEC annotated data improves performance significantly, as evidenced by a giant leap between the systems without GEC data (\textit{Korektor}, \textit{Synthetic trained}) and the systems trained on GEC data (\textit{AG finetuned}, \textit{GECCC finetuned} and \textit{Joint GEC+NMT}). Further addition of GEC data volume and domains is statistically significantly better ($p < 0.001$), as the only difference between \textit{AG finetuned} and \textit{\GECCC finetuned} systems is that the former uses the \textit{AKCES-GEC} corpus, while the latter is trained on larger and domain-richer \GECCC. Access to larger data and more domains in the multi-task setting is useful (compare \textit{Joint GEC+NMT} and \textit{AG finetuned} on newly added \textit{Natives Formal} and \textit{Natives Web Informal} domains), although direct training seems superior (\textit{GECCC finetuned} over \textit{Joint GEC+NMT}).

We further analyse the best model (\textit{GECCC finetuned}) and inspect its performance with respect to individual error types. For simpler analysis, we grouped all POS-related errors into two error types: \textit{POS} and \textit{POS:INFL} for words which are erroneous only in inflection and share the same lemma with their correction.

As we can see in Table~\ref{fig:model_error_analysis}, the model is very good at correcting local errors in diacritics (\textit{DIACR}), quotation (\textit{QUOTATION}), spelling (\textit{SPELL}) and casing (\textit{ORTH:CASING}). Unsurprisingly, small changes are easier than longer edits: similarly, the system is better in inflection corrections (\textit{POS:INFL}, words with the same lemma) than on \textit{POS} (correction involves finding a word with a different lemma).

Should the word be split or joined with an adjacent word, the model does so with a relatively high success rate (\textit{ORTH:WSPACE}). The model is also able to correctly reorder words (\textit{WO}), but here its recall is rather low. The model performs worst on errors categorized as \textit{OTHER}, which includes edits that often require rewriting larger pieces of text. Generally, the model has higher precision than recall, which suits the needs of standard GEC, where proposing a bad correction for a good text is worse than being inert to an existing error.

\begin{table}[t]
\small
\begin{tabular}{l|r|ccc}
\toprule
Error Type    & \# & P      & R      & $F_{0.5}$    \\
\midrule
\textit{DIACR} & 3\,617 & 86.84	& 88.77	& 87.22 \\
\textit{MORPH} & 610 & 73.58	& 55.91	& 69.20 \\
\textit{ORTH:CASING} & 1\,058 & 81.60	& 55.15	& 74.46 \\
\textit{ORTH:WSPACE} & 385 & 64.44	& 74.36	& 66.21 \\
\textit{OTHER} & 3\,719 & 23.59	& 20.04	& 22.78 \\
\textit{POS} & 2\,735 & 56.50	& 22.12	& 43.10 \\
\textit{POS:INFL} & 1\,276 & 74.47	& 48.22	& 67.16 \\
\textit{PUNCT} & 4\,709 & 71.42 & 61.17 & 69.10 \\
\textit{QUOTATION} & 223 & 89.44	& 61.06	& 81.83 \\
\textit{SPELL} & 1\,816 & 77.27	& 75.76	& 76.96 \\
\textit{WO} & 662 & 60.00	& 29.89	& 49.94 \\
\bottomrule
\end{tabular}
\caption{Analysis of \textit{GECCC finetuned} model performance on individual error types. For this analysis, all POS-error types were merged into a single error type POS.}
    \label{fig:model_error_analysis}
\end{table}

\section{Meta-evaluation of Metrics}
\label{sec:metrics}

There are several automatic metrics used for evaluating system performance on GEC dataset, although it is not clear which of them is preferable in terms of high correlation with human judgements on our dataset.

The most popular GEC metrics are the MaxMatch (M$^2$) scorer \cite{dahlmeier-ng-2012-better} and the ERRANT scorer \cite{bryant-etal-2017-automatic}.

The MaxMatch (M$^2$) scorer reports the F-score over the optimal phrasal alignment between a source sentence and a system hypothesis reaching the highest overlap with the gold standard annotation. It was used as the official metric for the CoNLL 2013 and 2014 Shared Tasks \cite{ng-etal-2013-conll,ng-etal-2014-conll} and is also used on various other datasets such as the German \textit{Falko-MERLIN GEC} \cite{boyd2018wnut} or Russian \textit{RULEC-GEC} \cite{rozovskaya-roth-2019-grammar}.

The ERRANT scorer was used as the official metric of the recent Building Educational Application 2019 Shared Task on GEC \cite{bryant-etal-2019-bea}. The ERRANT scorer also contains a set of rules operating over a set of linguistic annotations to construct the alignment and extract individual edits.

Other popular automatic metrics are the General Language Evaluation Understanding (GLEU) metric \cite{napoles-etal-2015-ground}, that additionally measures text fluency, and I-Measure \cite{felice-briscoe-2015-towards}, that calculates weighted accuracy of both error detection and correction.

\subsection{Human Judgements Annotation}

In order to evaluate the correlation of several GEC metrics with human judgements, we collected annotations of the original erroneous sentences, the manually corrected gold references and automatic corrections made by five GEC systems described in Section~\ref{sec:model}. We used the hybrid \textit{partial ranking with scalars} \cite{sakaguchi-van-durme-2018-efficient}, in which the annotators judged the sentences on a scale 0--10 (from ungrammatical to correct).\footnote{Recent works \cite{sakaguchi-van-durme-2018-efficient,novikova-etal-2018-rankme} both found partial ranking with scalars to be more reliable than direct assessment framework used by WMT~\cite{bojar-etal-2016-findings} and earlier GEC evaluation approaches \cite{grundkiewicz-etal-2015-human,napoles-etal-2015-ground}.} The sentences were evaluated with respect to the context of the document. In total, three annotators judged 1\,100 documents, sampled from the test set comprising about 4\,300 original sentences and about 15\,500 unique corrected variants and gold references of the sentences. The annotators annotated 127 documents jointly and the rest was annotated by a single annotator. This annotation process took about 170 hours. Together with the model training, data preparation and management of the annotation process, our rough estimation is about 300+ man-hours for the correlation analysis per corpus (language).

\subsection{Agreement in Human Judgements}
\label{sec:metric_judgement_aggrement}

For the agreement in human judgements, we report the \textit{Pearson correlation} and \textit{Spearman's rank correlation coefficient} between 3 human judgements of 5 automatic sentence corrections at the system- and sentence-level. At the \textit{sentence level}, the correlation of the judgements about the 5 sentence corrections is calculated for each sentence and each pair of the three annotators. The final sentence-level annotator agreement is the mean of these values over all sentences. 

At the \textit{system level}, the annotators' judgements for each system are averaged over the sentences, and the correlation of these averaged judgements is computed for each pair of the three annotators. In order to obtain smoother estimates (especially for Spearman's $\rho$), we utilize bootstrap resampling with 100 samples of a test set.

The human judgements agreement across domains is shown in Table~\ref{table:human_judgements_agreement}. On the sentence level, the human judgements correlation is high on the least erroneous domain \textit{Natives Formal}, implying that it is easier to judge the corrections in a low error density setting, and it is more difficult in high error density domains, such as \textit{Romani} and \textit{Second Learners} (compare error rates in Table~\ref{table:statistics_final_dataset}).

\begin{table}[t]
\small
\centering
\begin{tabular}{l|cc|cc}
\toprule
& \multicolumn{2}{c|}{Sentence level} & \multicolumn{2}{c}{System level} \\
Domain & $r$ & $\rho$ & $r$ & $\rho$ \\
\midrule
\textit{Natives Formal} & 87.13 & 88.76 & 92.01 & 92.52 \\
\textit{Natives Web Inf.} & 80.23 & 81.47 & 95.33 & 91.80 \\
\textit{Romani} & 86.57 & 86.57 & 88.73 & 85.90 \\
\textit{Second Learners} & 78.50 & 79.97 & 96.50 & 97.23 \\
\midrule
Whole Dataset & 79.07 & 80.40 & 96.11 & 95.54 \\
%\added[id=MS]{Domain Average} & 83.16 & 84.22 & 93.55 & 90.78 \\
\bottomrule
\end{tabular}
\caption{Human judgements agreement: Pearson ($r$) and Spearman ($\rho$) mean correlation between 3 human judgements of 5 sentence versions at sentence- and system-level.}
\label{table:human_judgements_agreement}
\end{table}

\begin{figure*}[!t]
    \includegraphics[width=0.5\hsize]{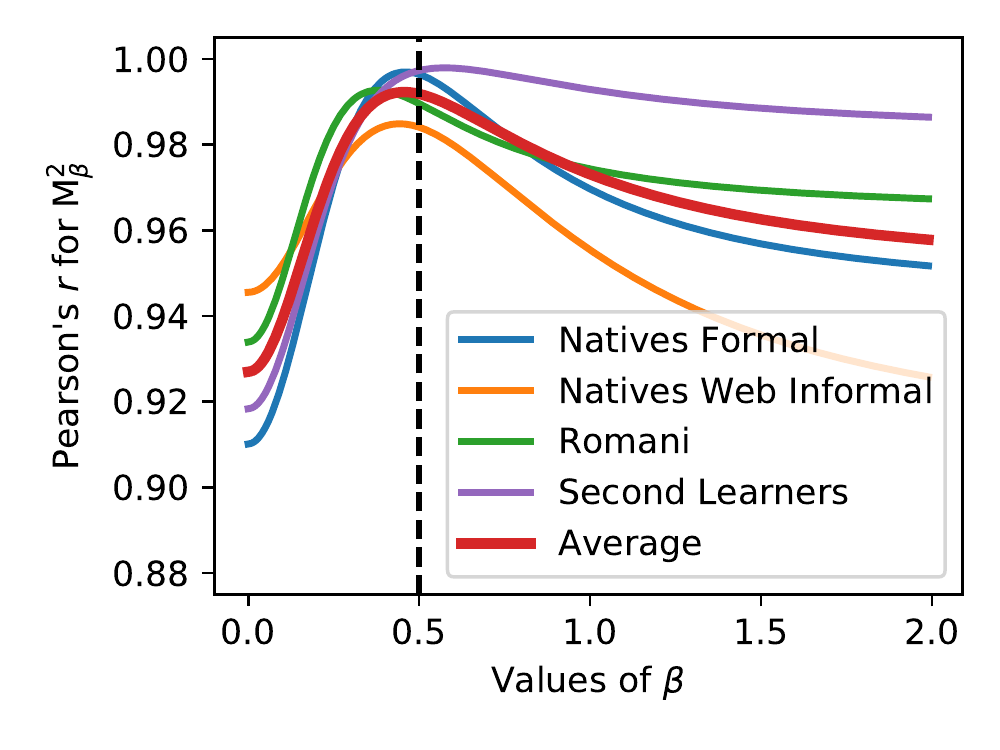}%
    \includegraphics[width=0.5\hsize]{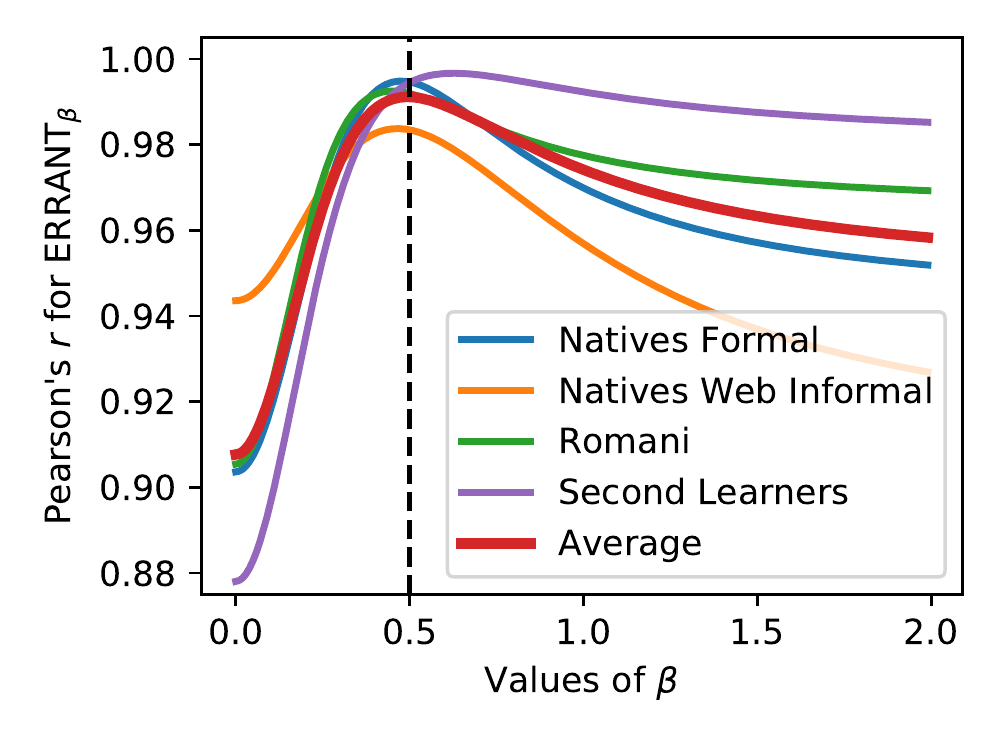}
    \caption{\textbf{Left:} System-level Pearson correlation coefficient $r$ between human annotation and M$^2_\beta$-scorer for various values of $\beta$. \textbf{Right:} The
    same correlation for ERRANT$_\beta$.}
    \label{fig:varying_beta_correlation}
\end{figure*}

\subsection{Metrics Correlations with Judgements}

Following \citet{napoles-etal-2019-enabling}, we provide a meta-evaluation of the following common GEC metrics robustness on our corpus:

\begin{citemize}
    \item MaxMatch (M$^2$) \cite{dahlmeier-ng-2012-better}
    \item ERRANT \cite{bryant-etal-2017-automatic}
    \item GLEU \cite{napoles-etal-2015-ground}
    \item I-measure \cite{felice-briscoe-2015-towards}
\end{citemize}

Moreover, we vary the proportion of recall and precision, ranging from $0$ to $2.0$ for M$^2$-scorer and ERRANT, as \citet{grundkiewicz-etal-2015-human} report that the standard choice of considering precision two times as important as recall may be sub-optimal.

\looseness1
While we considered both sentence-level and system-level evaluation in Section~\ref{sec:metric_judgement_aggrement},
the automatic metrics should by design be used on a whole corpus,
leaving us with only system-level evaluation. Given that the GEC systems perform differently on the individual domains (as indicated by Table~\ref{table:m2_human_scores}), we perform the correlation computation on each domain separately and report the average.

For a given domain and metric, we compute the correlation between the automatic metric evaluations of the five systems on one side
and the (average of) human judgements on the other side. In order to obtain a smoother estimate of Spearman's $\rho$ and also to estimate standard deviations, we employ bootstrap resampling again, with \hbox{100 samples}.

The results are presented in Table~\ref{tab:correlation_evaluation}. While
Spearman's $\rho$ has more straightforward interpretation, it also has a much higher
variance, because it harshly penalizes the differences in the ranking of systems with similar performance (namely \textit{AG finetuned} and \textit{Joint GEC+NMT} in our case). This fact
has previously been observed by \citet{machacek-bojar-2013-wmt}.

Therefore, we choose the most suitable GEC metric for our \GECCC dataset according to Pearson $r$, which implies that M$^2_{0.5}$ and ERRANT$_{0.5}$ are the metrics most correlating with human judgements. Of those two, we prefer the M$^2_{0.5}$ score, not due to its marginal superiority in correlation (Table~\ref{tab:correlation_evaluation}), but rather because it is much more language-agnostic compared to ERRANT, which requires a POS tagger, lemmatizer, morphological dictionary and language-specific rules.

% Original
% Our results confirm that both M$^2$-scorer and ERRANT with $\beta=0.5$ (chosen only by intuition for the CoNLL 2014 Shared task; \citealp{ng-etal-2014-conll}) correlate much better with human judgements, compared to $\beta=0.2$ and $\beta=1$. This is distinct from the results of \citet{grundkiewicz-etal-2015-human}, where $\beta=0.18$ correlates best, and also from the results of \citet{napoles-etal-2019-enabling}, where $\beta=0.2$ correlates better than $\beta=0.5$. The detailed plots of correlations of M$^2_\beta$ score and ERRANT$_\beta$ score with human judgements for $\beta$ ranging between $0$ and $2$, presented in Figure~\ref{fig:varying_beta_correlation}, show that optimal $\beta$ in our case lies between $0.4$ and $0.5$. However, we opt to employ the widely used $\beta=0.5$ because of its prevalence and because the difference to the optimal $\beta$ is marginal.

% MS: New version
Our results confirm that both M$^2$-scorer and ERRANT with $\beta=0.5$ (chosen only by intuition for the CoNLL 2014 Shared task; \citealp{ng-etal-2014-conll}) correlate much better with human judgements, compared to $\beta=0.2$ and $\beta=1$. The detailed plots of correlations of M$^2_\beta$ score and ERRANT$_\beta$ score with human judgements for $\beta$ ranging between $0$ and $2$, presented in Figure~\ref{fig:varying_beta_correlation}, show that optimal $\beta$ in our case lies between $0.4$ and $0.5$. However, we opt to employ the widely used $\beta=0.5$ because of its prevalence and because the difference to the optimal $\beta$ is marginal.

\begin{table}[t]
\centering
\begin{tabular}{l|cc}
\toprule
& \multicolumn{2}{c}{System level}  \\
Metric & $r$ & $\rho$ \\
\midrule
GLEU                            & 97.37 $\pm$ 1.52 & 92.28 $\pm$ 6.19 \\
I-measure                        & 95.37 $\pm$ 2.16 & 98.66 $\pm$ 3.21 \\
M$^2_{0.2}$      & 96.25 $\pm$ 1.71 & 93.27 $\pm$ 9.45 \\
M$^2_{0.5}$      & 98.28 $\pm$ 1.03 & 97.77 $\pm$ 4.27 \\
M$^2_{1.0}$        & 95.62 $\pm$ 1.81  & 93.22 $\pm$ 4.30 \\
ERRANT$_{0.2}$ & 94.66 $\pm$ 2.44 & 91.19 $\pm$ 4.76 \\
ERRANT$_{0.5}$ & 98.28 $\pm$ 1.04 & 98.35 $\pm$ 4.81 \\
ERRANT$_{1.0}$   & 95.70 $\pm$ 1.80 & 93.61 $\pm$ 4.47  \\
\bottomrule
\end{tabular}
\caption{System-level Pearson ($r$) and Spearman ($\rho$) correlation between the automatic metric scores and human annotations.}
\label{tab:correlation_evaluation}
\end{table}

Our results are distinct from the results of \citet{grundkiewicz-etal-2015-human}, where $\beta=0.18$ correlates best on the \textit{CoNLL 14 test set}. Nevertheless, \citet{napoles-etal-2019-enabling} demonstrate that $\beta=0.5$ correlates slightly better than $\beta=0.2$ on the \textit{FCE} dataset, but that $\beta=0.2$ correlates substantially better than $\beta=0.5$ on \textit{Wikipedia} and also on \textit{Yahoo} discussions (a dataset containing paragraphs of Yahoo! Answers, which are informal user answers to other users' questions).

In the latter work, \citet{napoles-etal-2019-enabling} propose that larger $\beta=0.5$ correlate better on datasets with higher error rate and vice versa, given that the \textit{FCE} dataset has 20.2\% token error rate, compared to the error rates of 9.8\% and 10.5\% of \textit{Wikipedia} and \textit{Yahoo}, respectively. The hypothesis seems to extend to our results and the results of \citet{grundkiewicz-etal-2015-human}, considering that the \textit{GECCC} dataset and the \textit{CoNLL 14 test set} have token error rates of 18.2\% and 8.2\%, respectively.

%MS: Commented out
%Still, we speculate that pure error rate is insufficient to predict suitable $\beta$, because the hypothesis does not extend to the individual \textit{domains} of \textit{GECCC} -- it suggests smaller $\beta$ for \textit{Natives Formal} with smaller error rate of 5.8\% and larger $\beta$ for \textit{Romani} with largest error rate of 26.2\%, which contradicts Figure~\ref{fig:varying_beta_correlation}.

\subsection{GEC Systems Results}

Table~\ref{table:m2_human_scores} presents both human scores for the GEC systems described in Section~\ref{sec:model} and also results obtained by the chosen M$^2_{0.5}$ metric. The results are presented both on the individual domains and the entire dataset.  Measuring over the entire dataset, human judgements and the $M^2$-scorer rank the systems in accordance. 

\looseness1
Judged by the human annotators, all systems are better than the ``do nothing'' baseline (the \textit{Original}) measured over the entire dataset, although \textit{Korektor} makes harmful changes in two domains: \textit{Natives Formal} and \textit{Natives Web Informal}. These two domains contain frequent named entities, which upon an eager change disturb the meaning of a sentence, leading to severe penalization by human annotators. \textit{Korektor} is also not capable of deleting, inserting, splitting or joining tokens. The fact that \textit{Korektor} sometimes performs detrimental changes cannot be revealed by the $M^2$-scorer as it assigns zero score to the \textit{Original} baseline and does not allow negative scores.

The human judgements confirm that there is still a large gap between the optimal \textit{Reference} score and the best performing models. Regarding the domains, the neural models in the finetuned mode that had access to data from all domains seemed to improve the results consistently across each domain. However, given the fact that the source sentences in the \textit{Second Learners} domain received the worst scores by human annotators, this domain seems to hold the greatest potential for future improvements.

\section{Conclusions}

We release a new Czech GEC corpus, the \GECCCLong (\GECCC). This large corpus with \nsentences sentences covers four diverse domains, including essays written by native students, informal website texts, essays written by Romani ethnic minority children and teenagers and essays written by non-native speakers. All domains are professionally annotated for GEC errors in a unified manner, and errors were automatically categorized with a Czech-specific version of ERRANT released at {\footnotesize\url{https://github.com/ufal/errant_czech}}. We compare several strong Czech GEC systems, and finally, we provide a meta-evaluation of common GEC metrics across domains in our data. We conclude that M$^2$ and ERRANT scores with $\beta=0.5$ are the measures most correlating with human judgements on our dataset, and we choose the M$^2_{0.5}$ as the preferred metric for the \GECCC dataset. 
The corpus is publicly available under the CC BY-SA 4.0 license at {\footnotesize\url{http://hdl.handle.net/11234/1-4639}}.

\section*{Acknowledgements}

This work has been supported by the Grant Agency of the Czech Republic, project EXPRO LUSyD (GX20-16819X). This research was also partially supported by SVV project number 260 575 and GAUK 578218 of the Charles University. The work described herein has been supported by and has been using language resources stored by the LINDAT/CLARIAH-CZ Research Infrastructure ({\footnotesize\url{https://lindat.cz}}) of the Ministry of Education, Youth and Sports of the Czech Republic (Project No. LM2018101). This work was supported by the European Regional Development Fund project “Creativity and Adaptability as Conditions of the Success of Europe in an Interrelated World” (reg. no.: CZ.02.1.01/0.0/0.0/16\_019/0000734).

We would also like to thank the reviewers and the TACL action editor for their thoughtful comments, which helped to improve this work.

\bibliography{tacl2018}
\bibliographystyle{acl_natbib}

\end{document}